\documentclass[letterpaper]{article} 
\usepackage{aaai25}  
\usepackage{times}  
\usepackage{helvet}  
\usepackage{courier}  
\usepackage[hyphens]{url}  
\usepackage{graphicx} 
\usepackage{natbib}  
\usepackage{caption} 
\usepackage{algorithm}
\usepackage{algorithmic}

\usepackage{amssymb}
\usepackage{booktabs}
\usepackage{tabularx}
\usepackage{multirow}
\usepackage{subfig}
\usepackage{amsmath}
\usepackage{makecell}

\usepackage{newfloat}
\usepackage{listings}
\DeclareCaptionStyle{ruled}{labelfont=normalfont,labelsep=colon,strut=off} 
\floatstyle{ruled}
\newfloat{listing}{tb}{lst}{}
\floatname{listing}{Listing}
\title{Explore In-Context Segmentation via Latent Diffusion Models}
\author{
    Chaoyang Wang\textsuperscript{\rm 1}, 
    Xiangtai Li\textsuperscript{\rm 2,3}\thanks{Project Leader}, 
    Henghui Ding\textsuperscript{\rm 4}, 
    Lu Qi\textsuperscript{\rm 5}, 
    Jiangning Zhang\textsuperscript{\rm 6}, \\
    Yunhai Tong\textsuperscript{\rm 1}, 
    Chen Change Loy\textsuperscript{\rm 3}, 
    Shuicheng Yan\textsuperscript{\rm 2,3}
}
\affiliations{
    \textsuperscript{\rm 1}School of Intelligence Science and Technology, Peking University, China\quad \\
    \textsuperscript{\rm 2}Skywork AI, Singapore \quad \\
    \textsuperscript{\rm 3}Nanyang Technological University, Singapore\quad \\
    \textsuperscript{\rm 4}Institute of Big Data, Fudan University, China\quad \\
    \textsuperscript{\rm 5}Wuhan University, China \quad \\
    \textsuperscript{\rm 6}Zhejiang University, China \\
    cywang@stu.pku.edu.cn, xiangtai94@gmail.com
}

\usepackage{bibentry}

\begin{document}

\maketitle


%

\begin{abstract}
    In-context segmentation has drawn increasing attention with the advent of vision foundation models. Its goal is to segment objects using given reference images.
    Most existing approaches adopt metric learning or masked image modeling to build the correlation between visual prompts and input image queries. 
    This work approaches the problem from a fresh perspective -- unlocking the capability of the latent diffusion model (LDM) for in-context segmentation and investigating different design choices.
    Specifically, we examine the problem from three angles: instruction extraction, output alignment, and meta-architectures. 
    We design a two-stage masking strategy to prevent interfering information from leaking into the instructions. 
    In addition, we propose an augmented pseudo-masking target to ensure the model predicts without forgetting the original images.
    Moreover, we build a new and fair in-context segmentation benchmark that covers both image and video datasets. 
    Experiments validate the effectiveness of our approach, demonstrating comparable or even stronger results than previous specialist or visual foundation models.
    We hope our work inspires others to rethink the unification of segmentation and generation. 
\end{abstract}

\begin{links}
    \link{Project page}{https://wang-chaoyang.github.io/project/refldmseg}
\end{links}

\begin{figure*}[t!]
\centering
\includegraphics[width=1.0\textwidth]{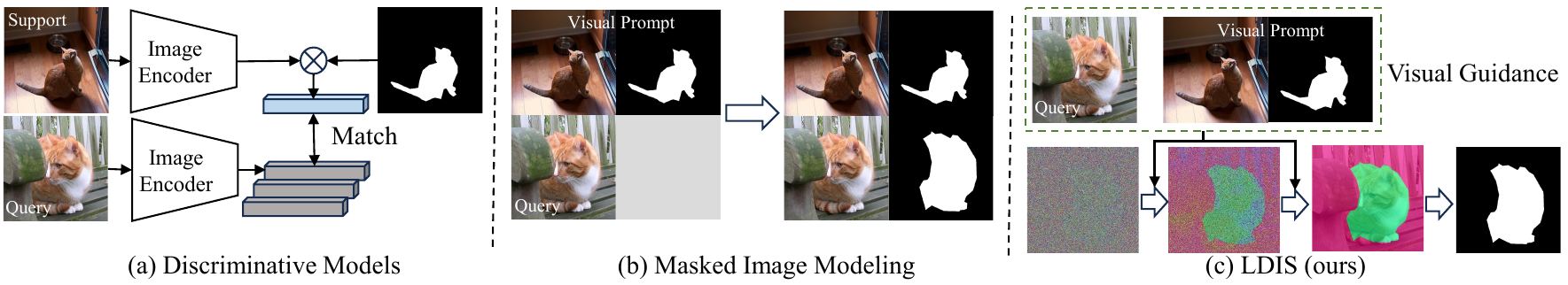}
\captionof{figure}{Method comparison. (a) Discriminative models match query images with support prototypes. (b) Masked image modeling methods adopt inpainting training. (c) Our LDM-based model generates segmentation masks guided by visual prompts. }
\label{fig:real_teasor}
\end{figure*}

\section{Introduction}
\label{sec:intro}

In-context learning~\cite{brown2020language,hummingbird,bar2022visual} provides a new perspective for cross-task modeling for vision and natural language processing (NLP). 
It enables a model to learn and predict according to the prompts. 
GPT-3~\cite{brown2020language} first introduced the concept of in-context learning, which refers to inferring solutions for unseen tasks by conditioning on input-output pairs provided as context. Subsequently, several studies~\cite{bar2022visual, Painter} explored in-context learning in the vision domain, where prompts are designed as visual task input-output pairs.

In the segmentation field~\cite{li2023transformer, MOSE, MeViS, GRES}, in-context learning serves a similar purpose to few-shot segmentation (FSS)~\cite{shaban2017one, ding2023self}. Most methods compute the matching distance between query images and support images, which act as visual prompts for in-context learning.

To address the strict constraints on data volume and category in FSS and enable generalization across different tasks, recent works~\cite{bar2022visual, Painter, SegGPT} have extended the concept to in-context segmentation, framing it as a mask generalization task (Fig.~\ref{fig:real_teasor}(b)).
This approach fundamentally differs from matching or prototype-based discriminative models (Fig.~\ref{fig:real_teasor}(a)), as it directly generates masks through mask decoding.
However, these methods typically require large datasets to learn such correspondences.

\begin{figure}[t!]
	\centering
	\includegraphics[width=1.0\linewidth]{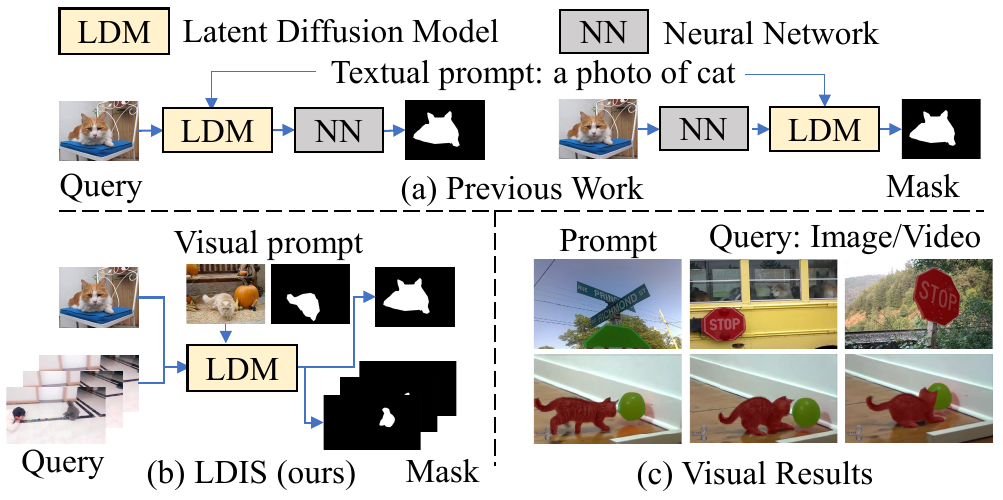}
	\caption{\textbf{Latent diffusion model for in-context segmentation.} (a) Previous works mainly rely on textual prompts and additional neural networks for segmentation. (b) Our proposed minimalist framework, LDIS. (c) Segmentation results on images and videos. 
 }
    \label{fig:teasear}
\end{figure}

Latent diffusion models (LDMs)~\cite{ho2020denoising, rombach2022high} have shown significant potential for generative tasks.
Several studies~\cite{zhang2023adding, rombach2022high} have demonstrated their strong performance in conditional image content creation.
Although LDMs were originally designed for generative tasks, there have been attempts to apply them to perceptual tasks as well. Fig.~\ref{fig:teasear}(a) shows the mainstream pipeline~\cite{zhao2023unleashing, xu2023open, geng2023instructdiffusion, baranchuk2022label} for LDM-based segmentation, which typically relies on textual prompts for semantic guidance and additional neural networks to support the LDM.
However, textual prompts are not always available in real-world scenarios, and relying on additional networks limits the exploration of LDM’s segmentation capabilities. This dependency can also degrade model performance when these auxiliary components are absent.
Based on this analysis, we propose that in-context segmentation can be reframed as a conditional image mask generation process, fully leveraging the generative potential of LDMs.

In this paper, we explore, for the first time, the potential of diffusion models for in-context segmentation, as illustrated in Fig.~\ref{fig:real_teasor}(c).
Our goal is to answer two key questions: First, can LDMs perform in-context segmentation? Second, what factors are crucial for performance, and how do they influence it?
To address these questions, we introduce the \textbf{L}atent \textbf{D}iffusion-based \textbf{I}n-context \textbf{S}egmentation framework, or \textbf{LDIS}, shown in Fig.~\ref{fig:teasear}(b). LDIS leverages visual prompts for guidance, eliminating the need for additional neural networks.
We focus our analysis on three critical factors: instruction extraction, output alignment, and meta-architectures.

First, we introduce a simple yet effective instruction extraction strategy. Experimental results show that these extracted instructions provide strong guidance, and our model remains robust even when instructions are incorrect.
Next, to bridge the gap between binary segmentation masks and 3-channel images, we design a novel output alignment target using pseudo-masking modeling.
We then propose two meta-architectures: LDIS-1 and LDIS-n. These differ in their input formulation, sampling steps, and optimization targets. Specifically, we design two optimization targets for LDIS-1, one in pixel space and the other in latent space. Experiments highlight the critical role of output alignment.
Unlike existing methods~\cite{SegGPT}, our approach focuses more on the architectural impact than on the size of the training data. We aim to use a dataset that is larger than typical few-shot datasets but significantly smaller than the datasets used for foundation models.
Finally, we introduce an in-context segmentation benchmark that covers image semantic segmentation, video object segmentation, and video semantic segmentation. We conduct extensive ablation studies and compare our method with previous works to demonstrate its effectiveness.
Our contributions are summarized as follows:

\begin{itemize}
    \item We unlock the in-context segmentation capabilities of latent diffusion models, enabling them to segment specified concepts using visual prompts alone, without relying on textual instructions or additional refinement networks.

    \item We investigate three key aspects, namely instruction extraction, output alignment, and meta-architectures, and highlighting the importance of accurate instructions, direct optimization targets,  and expressive power. 

    \item We propose an in-context learning benchmark, covering both image and video segmentation tasks, and show the effectiveness of our proposed LDIS on these tasks.
\end{itemize}

\section{Related Work}

\noindent
\textbf{Diffusion Model Design.} Diffusion models~\cite{ho2020denoising,song2020denoising,song2020score} have shown remarkable performance on generation tasks, such as image generation~\cite{rombach2022high,zhang2023adding,ho2022classifier,dhariwal2021diffusion}, image editing~\cite{brooks2023instructpix2pix,lugmayr2022repaint,saharia2022palette,meng2021sdedit,hertz2022prompt,li2022videoknet}, image super resolution~\cite{saharia2022image,ho2022cascaded}, video generation~\cite{harvey2022flexible,yang2023diffusion}, and point cloud~\cite{luo2021diffusion,zeng2022lion}. 
Although the diffusion model is initially designed for generation tasks, several works employ it for segmentation through two pipelines. The first pipeline treats the diffusion model as a feature extractor. These works~\cite{li2023sd4match,zhao2023unleashing,geng2023instructdiffusion,xu2023open,xie2023mosaicfusion,baranchuk2022label,khosravi2023few,li2023grounded,wan2023harnessing} typically rely on a decoder head for post-processing. Conversely, the second pipeline~\cite{chen2023diffusiondet,chen2023generalist,DiffusionInst,amit2021segdiff,le2023maskdiff} extracts features through a pre-trained backbone, then employs the diffusion model as the decoder head. 
Recently, several works~\cite{qi2024unigs} also explore using LDM to generate segmentation masks. 
However, these studies mainly focus on class-agnostic mask generation with no reference object as the context.
These additional neural networks greatly influence our judgment of the true capabilities of diffusion model \textit{itself} in the segmentation task. 
Moreover, many of these works require textual prompts for guidance. However, the textual prompts, such as categories or captions, are not always available in real-world scenarios.

\noindent
\textbf{In-context Learning.}
GPT-3~\cite{brown2020language} firstly defines in-context learning, which is interpreted as inferring on unseen tasks conditioning on some input-output pairs given as contexts, also known as prompts.  
As a new concept in computer vision, in-context learning motivates several attempts~\cite{alayrac2022flamingo,bar2022visual,lu2022unified,Painter,SegGPT,hummingbird,zhang2023makes}.
The work~\cite{bar2022visual} is the first to adopt masked image modeling (MIM)~\cite{bao2021beit,he2022masked,xie2022simmim} as a visual in-context framework. 
Painter~\cite{Painter} and SegGPT~\cite{SegGPT} follow the same spirit but scale up with massive training data.
Different from MIM, we aim to explore in-context segmentation with the latent diffusion model to explore the potential of condition generation, where we are the first to carry out this study.

\noindent
\textbf{Few-shot Segmentation.} This task aims to segment query images given support samples. 
The current works~\cite{wang2019panet,tian2020prior,yang2020prototype,liu2022learning,liu2022dynamic} typically draw on the idea of metric learning by matching spatial location features with semantic centroids. Furthermore, two-branch conditional networks~\cite{shaban2017one,Rakelly2018ConditionalNF,lu2021simpler,he2023prototype}, 4D dense convolution~\cite{hong2022cost,min2021hypercorrelation} and transformer-based architecture~\cite{zhang2022feature,shi2022dense,xie2021few,kim2023universal} are also widely adopted by researchers. Although few-shot segmentation and in-context segmentation share a similar episode paradigm, the dataset used in few-shot segmentation is typically very small, making the model prone to over-fitting, which may influence the evaluation of the generalization ability.

\noindent
\textbf{Parameter Efficient Tuning.} 
Research works in this domain~\cite{houlsby2019parameter,li2021prefix,hu2021lora,zaken2021bitfit,guo2020parameter} aim to fine-tune only a tiny portion of parameters to adapt the pre-trained foundation models to various downstream tasks. 
They maintain the pre-trained knowledge of the foundation models. 
However, they suffer from inadequate expressive power. 
In our experiments, we adopt the low-rank adaptation (LoRA)~\cite{hu2021lora}, a tool widely used in diffusion models, to demonstrate the task gap between generation and segmentation. 
Although some works~\cite{bahng2022exploring,gal2022image,khani2023slime} try to avoid the dilemma by fine-tuning the prompts only, it is time-costly to learn and restore a new embedding for each prompt. In contrast, we employ a prompt encoder to extract in-context instructions from prompts.

\section{Method}
\label{sec:method}

This section first introduces the preliminaries of diffusion models and in-context segmentation, then analyzes the design of LDIS from three aspects, namely instruction extraction, output alignment, and meta-architecture, respectively. The notations are illustrated in Tab.~\ref{tab:notation}.

\begin{figure*}[t!]
	\centering
	\includegraphics[width=0.91\linewidth]{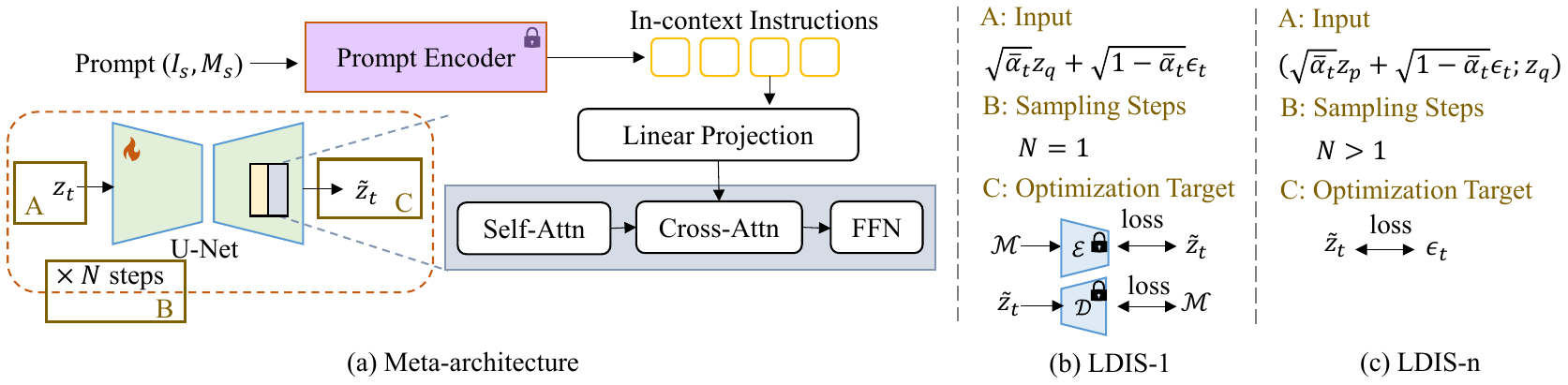}
	\caption{\textbf{Our proposed LDIS}. \textbf{Left:} Meta-architecture. Our model operates as a minimalist and generates the mask under the guidance of in-context instructions. \textbf{Right:} The two variants of our meta-architecture differ in input formulation, sampling steps, and optimization target. Notations are illustrated in Tab.~\ref{tab:notation}.}
    \label{fig:method_train}
\end{figure*}

\subsection{Preliminaries}
\label{subsec:prelim}
\noindent
\textbf{Diffusion Model.} 
Diffusion models belong to probabilistic generative models that define a chain of forward and backward processes. 
In the forward process, the model gradually corrupts the data sample $z_0$ into a noisy latent $z_t$ for $t\in \{1...T\}$:
$q(z_t|z_0)=\mathcal{N}(z_t;\sqrt{\overline{\alpha}_t}z_0,(1-\overline{\alpha}_t)I)$
, where $\overline{\alpha}_t=\prod _{i=0}^T\alpha_s=\prod_{i=0}^T(1-\beta_s)$ and $\beta$ is determined by noise scheduler. 
During training, the model learns to predict the noise $\epsilon_\theta(z_t,t)$ under the supervision of $L_2$ loss as
$\mathcal{L}=\frac{1}{2}||\epsilon_\theta(z_t,t)-\epsilon (t)||^2$.
During inference, the model starts from a random noise $z_T~\sim \mathcal{N}(0,1)$, gradually predicting the noise. 
Then, it reconstructs the original data $z_0$ with $T$ steps based on the estimated noise.

\noindent
\textbf{In-context Segmentation.}
Define query image $I_q$ and context set $S={(I_i, M_i)}_{i=1}^K$, where $I_i$ is a prompt image belonging to a specific visual concept, $M_i$ is the corresponding mask and $K$ is the number of prompts. The model aims to learn a segmentation function $g(I_q,S) \mapsto M_q$ such that based on the context information, it accurately segments target regions in the query image that are similar to the context.

\begin{table}[t]
\centering
\small
\resizebox{\linewidth}{!}{
\begin{tabular}{cc|cc}
\toprule[0.1em]
Notation & Definition & Notation & Definition \\
\midrule[0.1em]
$\mathcal{E}$ & VAE Encoder & $\mathcal{D}$ & VAE Decoder \\
$E_\tau$ & Prompt Encoder & $\mathcal{M}$ & Pseudo Mask \\
$M_q$ & Query Mask & $I_q$ & Query Image \\
$M_s$ & Prompt Mask & $I_s$ & Prompt Image \\
$M$ & Ground Truth & $\tau$ & Instruction \\
$z$ & Latent Input & $\tilde{z}$ & Latent Prediction\\
$z_q$ & Latent Query & $z_p$ & Latent Pseudo Mask \\
\bottomrule[0.1em]
\end{tabular}
}
\caption{Illustration of some notations in the Method section.}
\label{tab:notation}
\end{table}

\subsection{Framework}
\label{subsec:framework}

The LDM is orinally designed for generative tasks.
Most existing works~\cite{geng2023instructdiffusion, zhao2023unleashing} that apply LDMs to segmentation rely on task-specific decoders to process intermediate features or refine imperfect segmentation results.
However, incorporating such decoders limits the generative potential of LDMs.
To address this, we use Stable Diffusion (SD) as the base model with minimal modifications to fully explore its capabilities.
In this subsection, we investigate three key factors that influence the process: instruction extraction, output alignment, and meta-architectures.

\noindent
\textbf{Instruction Extraction.}
Instructions play an essential role in LDM. They act as the compressed prompt representation and guide the generation process.
To align with SD, it is intuitive to employ the CLIP vision encoder as the prompt encoder and use a binary mask to filter foreground information. However, this simple approach brings about the leakage of interference. Taking CLIP ViT L/14 as an example, it takes as input a $224\times224$ image $I_s$ and down-samples it to $16\times16$. In this process, the information of background or irrelevant targets is distributed among 196 tokens. Moreover, the down-sampled binary mask $M_s$ is ambiguous and cannot precisely represent the boundary of targets. 
To this end, we propose a two-stage masking strategy consisting of pre-masking and post-masking. In the pre-masking stage, the mask $M_s$ is taken as inputs along with the prompt image $I_s$, described as Eq.~\ref{equ:instruction}. 
\begin{equation}
\label{equ:instruction}
\tau = F\left(E_\tau\left(I_s, M_s\right)\right), 
\end{equation} 
where $F$ indicates the linear projection for alignment.
The binary mask $M_s$ filters the foreground tokens in this post-masking stage. In practice, we employ it as an attention map in cross-attention layers.

\noindent
\textbf{Output Alignment.}
We employ LDM for the segmentation task, so the inconsistency between 1-channel masks and 3-channel images is non-negligible.
A pseudo mask must be designed to align the gaps as an intermediate step toward the binary segmentation mask. 

An intuitive method follows a mapping rule that transforms the binary masks $M$ to 3-channel pseudo masks $\mathcal{M}_{v}$:
\begin{equation}
  \mathcal{M}_{vi}=\left\{ \begin{array}{l}
	\left( b,a,(a+b)/2 \right) ,\ M_i\in bg  \\
	\left( a,b,(a+b)/2 \right) ,\ M_i\in fg \\
    \end{array} \right.   ,
\label{equ:pm1}
\end{equation}
where $bg$ and $fg$ indicate background and foreground, respectively. $M_i$ is the value in position $i$. $a$, $b$ are both scalar, indicating the value of a specific channel in the pseudo masks. We set $a<b$.   

In the inference stage, the binary segmentation mask can be recovered with simple arithmetic operations:
\begin{equation}
    \label{equ:pm1_infer}
    \tilde{M}=
	\tilde{\mathcal{M}}_{v}[1]>\tilde{\mathcal{M}}_{v}[0]. 
\end{equation}
where $\tilde{M}$ and $\tilde{\mathcal{M}}_{v}$ indicate the predicted segmentation mask and vanilla pseudo masks, respectively. 
$[k]$ means the value in the $k^{th}$ channel.

Beyond the vanilla design, we also propose an augmented strategy to fuse the information of images into pseudo masks. Denote the query image as $I_q$, and the augmented pseudo masks are formulated as follows: 
\begin{equation}
    \label{equ:pm2}
    \mathcal{M}_a = (1-\gamma)\mathcal{M}_v+\gamma I_q, 
\end{equation}
where $\gamma$ controls the strength of the information of the image.

\noindent
\textbf{Meta-architectures.}
As shown in Fig.~\ref{fig:method_train}, we explore two representative meta-architectures, namely LDIS-1 and LDIS-n. They mainly differ in the input formats, sampling steps, and optimization targets. 

\noindent
\textbf{\emph{LDIS-1}} indicates \textbf{one-step sampling} and the optimization target is the segmentation mask itself.
As shown in Fig.~\ref{fig:method_train}(a), a noise variant $\epsilon_t$ is added to the latent variant $z_q$. 
The model takes as input the noisy latent $z_t = \sqrt{\overline{\alpha}_t}z_q+\sqrt{1-\overline{\alpha}_t}\epsilon_t$, where $\epsilon_t$ is the output of the noise scheduler, $t$ controls the noise strength. 

We propose two optimization strategies that align the model outputs with the ground truth in pixel space or latent space, respectively. 
We employ the L2 loss, typically used in LDM, rather than any explicit segmentation loss.

\begin{equation}
    \label{equ:loss_feat1}
    \mathcal{L}_{fp} =\mathbb{E}_{z_t,\tau}\left[ ||\mathcal{M}-\tilde{\mathcal{M}}_t||_{2}^{2} \right], \\
\end{equation}

\begin{equation}
    \label{equ:loss_feat1_latent}
    \mathcal{L}_{fl} =\mathbb{E}_{z_t,\tau}\left[ ||z_{p}-\tilde{z}_t||_{2}^{2} \right],
\end{equation}
where $\tilde{\mathcal{M}}_t = \mathcal{D}(\tilde{z}_t)$ and $z_{p}=\mathcal{E}(\mathcal{M})$.

In the inference stage, LDIS-1 conducts only one-time step and outputs the segmentation (pseudo) masks. The video is treated as a sequence of images. 
The first frame and its annotation are used as prompts, and subsequent frames are inferred conditioned on it. 
For videos containing multiple categories, we first calculate the probability of each category as a foreground in turn and then select the category with the highest probability:
\begin{equation}
\begin{aligned}
    \label{equ:video_infer1}
    \tilde{p}_c = \frac{\exp(\mathcal{M}[1])}{\exp(\mathcal{M}[0])}, \quad
    p_c = \frac{\tilde{p}_c}{1+\sum_{i=1}^C{\tilde{p}_i}},
\end{aligned}
\end{equation}
where $\tilde{p}_c$ indicates the normalized foreground probability map for category $c$. $C$ is the number of categories. $\mathcal{M}[i]$ means the value in channel $i$ of pseudo masks. It is evident that $\tilde{p}_0$, as the background's probability, equals $1$.

\noindent
\textbf{\emph{LDIS-n}} indicates \textbf{multi-step sampling} and employs an indirect optimization strategy. Unlike LDIS-1, it starts from Gaussian noise and gradually denoises to get the final segmentation mask.  

A plain SD architecture is not suitable for LDIS-n. We make minimal but necessary modifications to the architecture by extending the input dimension from 4 to 8. 
Specifically, denote the latent expression of query image and pseudo mask as $z_q\in \mathbb{R}^{4\times H\times W}$ and $z_{p} \in \mathbb{R}^{4\times H\times W}$, we get $z_t \in \mathbb{R}^{8\times H\times W}$ by concatenating the noisy pseudo mask latent with $z_q$ in the channel dimension. The noisy pseudo mask latent is obtained by adding noise $\epsilon_t$ to $z_{p}$:
\begin{equation}
    \label{equ:p2p_train}
    z_t=\mathrm{CONCAT} ((\sqrt{\overline{\alpha}_t}z_{p}+\sqrt{1-\overline{\alpha}_t}\epsilon_t);z_q),
\end{equation}
where the noise scheduler determines $t$.

Similar to LDIS-1, the ICS model $f$ inputs the latent variable $z_t$ and the instructions $\tau$, but outputs the estimation of the noise rather than the pseudo mask. We also adopt L2 loss as follows:
\begin{equation}
    \label{equ:loss_noise} \mathcal{L}_n=\mathbb{E}_{z_t,t,\tau}\left[ ||\epsilon_t-\tilde{z_t}||_{2}^{2} \right]. 
\end{equation}

To reduce the randomness caused by initial noise, enhance the influence of in-context instructions, and ensure consistency between outputs and queries, we employ classifier-free guidance (CFG)~\cite{ho2022classifier}. The query latent $z_q$ and condition $\tau$ are randomly set to null embedding with probability $p=0.05$ in the training stage.

We also adopt CFG in the inference stage. Specifically, LDIS-n outputs the $\tilde{z}_t(z_q,\tau)$ on the basis of three conditional outputs $\tilde{z}_t(z_q,\tau)$, $\tilde{z}_t(\varnothing,\varnothing)$, $\tilde{z}_t(z_q,\varnothing)$ (Eq.~\ref{equ:cfg_infer}). 
\begin{equation}
\begin{aligned}
    \label{equ:cfg_infer}
    \tilde{z}_t(z_q,\tau) &= \tilde{z}_t(\varnothing,\varnothing) \\
    &+ \gamma_q\cdot \left( \tilde{z}_t(z_q,\varnothing)-\tilde{z}_t(\varnothing,\varnothing)\right) \\
    &+ \gamma_\tau \cdot \left( \tilde{z}_t(z_q,\tau)-\tilde{z}_t(z_q,\varnothing)\right),
\end{aligned}
\end{equation}
where $\gamma_q$ and $\gamma_\tau$ control the guidance of query and in-context instruction, respectively.

\section{Experiments}
\label{sec:exp}

We begin by outlining the experimental settings, followed by a comprehensive comparison of our methods with existing specialist and generalist models. Finally, we perform ablation studies to evaluate the effectiveness of our design choices.

\subsection{Experimental Setting}
\label{subsec:expsetup}

\begin{table}[t]
\setlength{\tabcolsep}{1.7pt}
\centering
\begin{tabular}{clccccc}
\toprule[0.1em]
\multirow{2}{*}{Dataset} & \multirow{2}{*}{Task} & \multirow{2}{*}{\#Category} & \multicolumn{2}{c}{\#Videos} & \multicolumn{2}{c}{\#Images} \\ \cline{4-7} 
 &  &  & Train & Val & Train & Val \\ \hline
PASCAL & ISS & 20 & - & - & 10582 & 2000 \\
COCO & ISS & 80 & - & - & 82081 & 5000 \\
DAVIS-16 & VOS & - & 30 & 16 & (2064) & - \\
VSPW & VSS & 58 & 1000 & 100 & (16473) & - \\ 
\toprule[0.1em]
\end{tabular}
\caption{\textbf{Details of the combined datasets.} We choose two image semantic segmentation (ISS) datasets, one video object segmentation (VOS) dataset, and one video semantic segmentation (VSS) dataset. (N) means the equivalent number of images.}
\label{tab:trainrule}
\end{table}

\noindent
\textbf{Benchmark Details.}
The in-context segmentation model aims to solve multiple tasks with \textit{one} model, regardless of data type and domain. To this end, we adopt several popular datasets as part of our benchmark (Tab.~\ref{tab:trainrule}), including PASCAL~\cite{everingham2010pascal}, COCO~\cite{lin2014microsoft}, DAVIS-16~\cite{perazzi2016benchmark} and VSPW~\cite{miao2021vspw}. The training data for VSPW is sampled every four frames. All `stuff' categories are annotated as background.

\noindent
\textbf{Implementation Details.} 
We utilize the SD 1.5 model as the initialization and set the resolution as $256\times 256$. Our model is jointly trained on the combined dataset for 160K iterations with a batch size of 64. We employ an AdamW optimizer. Alpha CLIP~\cite{sun2023alphaclip} ViT-L is adopted as the prompt encoder. 
We set the CFG coefficient for query and instructions as 1.5 and 7, respectively. 
Our training follows the spirit of episodic learning. For the image dataset, images with the same semantic labels are considered a pair of queries and prompts. The video dataset follows the image dataset but requires that the query and prompt come from the same video.

\noindent
\textbf{Evaluation Metrics.}
%
We adopt the class mean intersection over union as our evaluation metric in empirical study, which is formulated as $\mathrm{mIoU}=\frac{1}{C}\sum_{i=1}^{C}\mathrm{IoU}_i$. $C$ is the number of classes except background. We also report the foreground-background IoU for other image-level tasks in our benchmark following ~\cite{tian2020prior}. For video-level tasks, we respectively adopt their evaluation metrics in DAVIS-16~\cite{perazzi2016benchmark} and VSPW~\cite{miao2021vspw}.

\subsection{Benchmark Results}

\begin{table*}[!t]
\small
\setlength{\tabcolsep}{2.5pt}
\centering
\begin{tabular}{clcccccccc}
\toprule[0.1em]
\multirow{2}{*}{Modeling} & \multirow{2}{*}{Method} & \multicolumn{1}{c|}{Backbone} & \multicolumn{2}{c}{PASCAL} & \multicolumn{2}{c|}{COCO} & \multicolumn{1}{c}{DAVIS-16} & \multicolumn{2}{c}{VSPW} \\ 
& & \multicolumn{1}{c|}{/ Initialization} & mIoU & \multicolumn{1}{c}{FB-IoU} & mIoU & \multicolumn{1}{c|}{FB-IoU} &  \multicolumn{1}{c}{$\mathcal{J}$\&$\mathcal{F}$} & mIoU & fwIoU \\ 
\toprule[0.1em]
\multirow{5}{*}{\makecell[c]{Discriminative\\Modeling}} & \multicolumn{1}{l}{PFENet~\cite{tian2020prior}} & \multicolumn{1}{c|}{R50} & 76.4 & \multicolumn{1}{c}{88.1} & 49.4 & \multicolumn{1}{c|}{76.9} & - & - & - \\
& \multicolumn{1}{l}{SVF~\cite{sun2022singular}}& \multicolumn{1}{c|}{R50} & 77.0 & \multicolumn{1}{c}{88.5} & 49.2 & \multicolumn{1}{c|}{77.0} & - & - & - \\
& \multicolumn{1}{l}{VTM~\cite{kim2023universal}}& \multicolumn{1}{c|}{R50} & 73.9 & \multicolumn{1}{c}{85.4} & 52.1 & \multicolumn{1}{c|}{76.2} & - & - & - \\ 
& \multicolumn{1}{l}{DCAMA~\cite{shi2022dense}}& \multicolumn{1}{c|}{R50} & 71.4 & \multicolumn{1}{c}{85.2} & 43.5 & \multicolumn{1}{c|}{73.2} & - & - & - \\ 
& \multicolumn{1}{l}{VPD~\cite{zhao2023unleashing}}& \multicolumn{1}{c|}{SD 1.5} & 75.3 & \multicolumn{1}{c}{86.9} & 48.6 & \multicolumn{1}{c|}{75.6} & - & - & - \\
 \hline
\multirow{3}{*}{MIM} & \multicolumn{1}{l}{Painter~\cite{Painter}}& \multicolumn{1}{c|}{ViT-L} & 53.7 & \multicolumn{1}{c}{69.7} & 30.8 & \multicolumn{1}{c|}{58.3} & \multicolumn{1}{c}{76.7} & 12.2 & 76.7 \\
& \multicolumn{1}{l}{SegGPT~\cite{SegGPT}}& \multicolumn{1}{c|}{Painter} & 75.6 & \multicolumn{1}{c}{86.7} & 50.7 & \multicolumn{1}{c|}{76.9} & \multicolumn{1}{c}{\textbf{80.5}} & \textbf{61.7} & \textbf{93.3} \\ 
& \multicolumn{1}{l}{SegGPT$^\dag$~\cite{SegGPT}}& \multicolumn{1}{c|}{Painter} & 65.2 & \multicolumn{1}{c}{81.4} & 39.5 & \multicolumn{1}{c|}{71.2} & \multicolumn{1}{c}{73.4} & 50.4 & 90.0 \\ \hline
SAM & \multicolumn{1}{l}{PerSAM~\cite{zhang2023personalize}} & \multicolumn{1}{c|}{SAM} & 47.6 & \multicolumn{1}{c}{69.4} & 25.5 & \multicolumn{1}{c|}{57.5} & \multicolumn{1}{c}{68.7} & 42.6 & 79.1 \\ 
 \hline
\multirow{3}{*}{LDM} & \multicolumn{1}{l}{Prompt Diffusion~\cite{wang2023context}}& \multicolumn{1}{c|}{SD 1.5} & 9.0 & \multicolumn{1}{c}{40.1} & 5.9 & \multicolumn{1}{c|}{40.7} & \multicolumn{1}{c}{-} & - & - \\
& \multicolumn{1}{l}{LDIS-n (ours)}& \multicolumn{1}{c|}{SD 1.5} & 76.7 & \multicolumn{1}{c}{87.2} & 52.6 & \multicolumn{1}{c|}{76.3} & \multicolumn{1}{c}{64.7} & 30.5 & 79.7 \\ 
& \multicolumn{1}{l}{LDIS-1 (ours)}& \multicolumn{1}{c|}{SD 1.5} & \textbf{85.3} & \multicolumn{1}{c}{\textbf{93.1}} & \textbf{62.6} & \multicolumn{1}{c|}{\textbf{84.5}} & \multicolumn{1}{c}{67.8} & 41.6 & 87.8 \\ 

\toprule[0.1em]
\end{tabular}
\caption{\textbf{Benchmark results.} 
We compare our method with several representative specialist models and generalist models. The \textbf{bold} entries represent
the best performance. $^\dag$ indicates the number of reproductions with a resolution of 256.}
\label{tab:benchmark}
\end{table*}

\noindent
\textbf{Baselines.}
We report specialist models and generalist models as baselines. We take PFENet, SVF, VTM, DCAMA, and VPD as specialist baselines. Specifically, the first four methods are proposed for few-shot tasks, which aligns closely with our setting. 
VPD exploits the features extracted by SD UNet.
All the specialist models are jointly trained on the combined datasets. 
As for generalist models, we take Painter~\cite{Painter}, SegGPT, PerSAM~\cite{zhang2023personalize} and Prompt Diffusion~\cite{wang2023context} into comparisons. 
They represent several classic modeling approaches, respectively.
In detail, Painter and SegGPT are based on masked image modeling.
Prompt Diffusion adopts a ControlNet-like architecture. 
PerSAM leverages the robust segmentation capabilities from SAM~\cite{kirillov2023segment}.

\noindent
\textbf{Quantitative Results.}
We make comprehensive comparisons in Tab.~\ref{tab:benchmark}.
For image tasks, LDIS-1 achieves the best results compared with these methods. LDIS-n also achieves decent performance of 76.7 mIoU on PASCAL and 52.6 mIoU on COCO. 
Our approaches perform comparablely to the generalist models for video tasks. Two factors explain this phenomenon.
One is the data volume. The training data of the other generalist models is far beyond ours. The other is the spatial prior --- the content of the first and subsequent frames is highly similar in video segmentation tasks. The other generalist models use spatial information by taking the pixel-level prompts as inputs. However, our model only relies on the visual embedding extracted by the prompt encoder without spatial knowledge. Achieving state-of-the-art on all metrics is not the primary goal of this work, and we leave it for future work.

\begin{figure*}[ht]
	\centering
	\includegraphics[width=0.85\linewidth]{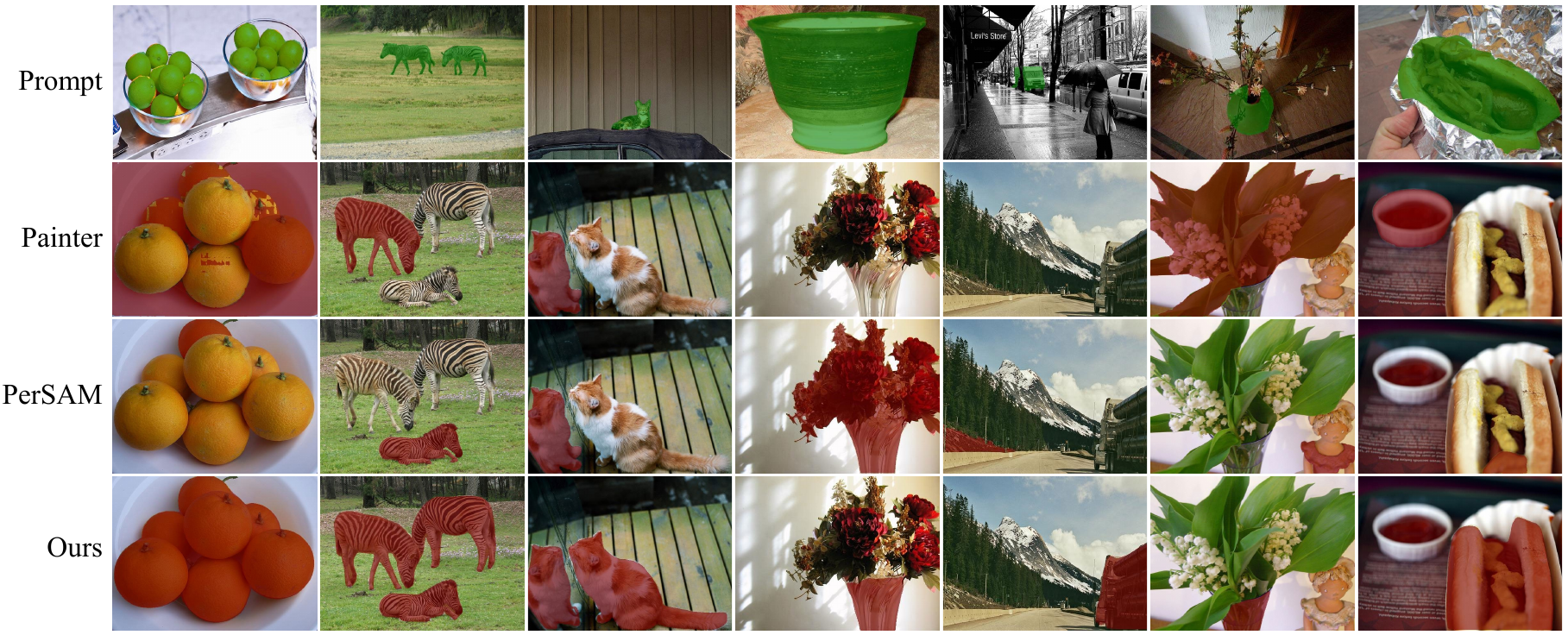}
	\caption{Visualization of segmentation results. We compare our LDIS-1 with Painter~\cite{Painter} and PerSAM~\cite{zhang2023personalize} on the COCO dataset.
 }
\label{fig:vis_coco}
\end{figure*}

\noindent
\textbf{Qualitative Results.}
Fig.~\ref{fig:vis_coco} shows the visual comparison between our model and previous works on the COCO dataset. It is evident from these results that our model successfully segments the target region. In contrast, other methods fail to establish the semantic connection between prompt and query, leading to missing and false-positive predictions.

\begin{figure*}[!ht]
	\centering
	\includegraphics[width=0.85\linewidth]{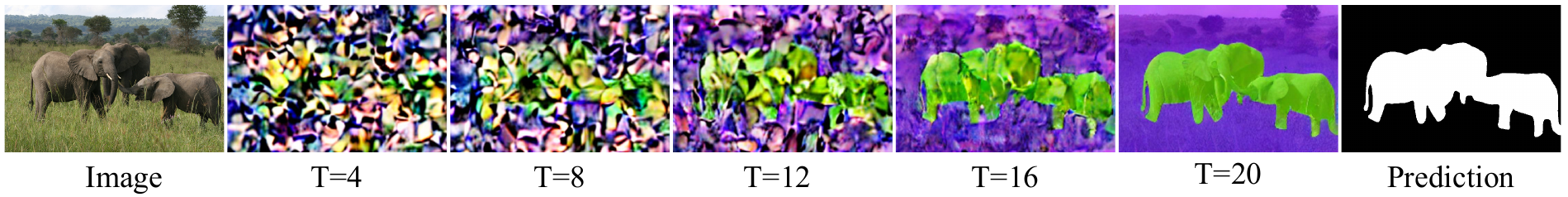}
	\caption{\textbf{Visualizations at different time steps.} LDIS-n captures low-frequency components at the beginning and then generates high-frequency information as the denoising process approaches completion. The number of denoising steps is 20.}
	\label{fig:pixlatent}
\end{figure*}

\subsection{Ablation Study and More Analysis}

In this subsection, we conduct comprehensive ablation experiments to study the subtle effects of different designs. COCO dataset with fixed prompt-query pairs is used for evaluation. The experiments are conducted from three aspects, namely instruction extraction, output alignment, and meta-architectures, respectively.

\noindent
\textbf{Instruction Extraction.} It is crucial to provide accurate visual instructions to guide the model in segmenting the specified concepts. However, the quality of instructions is affected by three factors. The initial factor to be considered is the number of visual prompts. As shown in Tab.~\ref{tab:ab_kshot}, the performance of LDIS-n first grows significantly as the number of visual prompts increases from 1 to 5 and then saturates around 10.
Multiple visual prompts provide different perspectives on the same concept, leading to more accurate estimations. However, too many prompts may cause information interference.
The second factor is how instructions are extracted. Tab.~\ref{tab:ab_alpclip} provides the ablation results of a two-stage masking strategy. Instructions without pre-masking or post-masking bring about unnecessary information leakage, and these models only achieve 43.8 and 49.7 mIoU. Models with a two-stage masking strategy get the best performance of 52.6 mIoU.

\begin{table}[htbp]
\centering
\begin{tabular}{cccccc}
\toprule[0.1em]
Number of Prompt & 1 & 3 & 5 & 10 & 20\\
\midrule[0.1em]
mIoU & 49.7 & 55.3 & 56.2 & 56.7 & 55.7 \\
\bottomrule[0.1em]
\end{tabular}
\caption{LDIS-n with different number of prompts.}
\label{tab:ab_kshot}
\end{table}

\begin{table}[ht]
\footnotesize
\begin{minipage}{0.48\linewidth}
\centering
\setlength{\tabcolsep}{2.0pt}
\begin{tabular}{cccc}
\toprule[0.1em]
Pre-Mask & Post-Mask & mIoU \\
\midrule[0.1em]
\checkmark & - & 43.8 \\
- & \checkmark & 49.7 \\
\checkmark & \checkmark & 52.6 \\
\bottomrule[0.1em]
\end{tabular}
\caption{Instruction extraction strategy on LDIS-n.}
\label{tab:ab_alpclip}
\end{minipage}
\hfill
\begin{minipage}{0.48\linewidth}
\centering
\begin{tabular}{cc}
\toprule[0.1em]
Method & mIoU \\
\midrule[0.1em]
$\mathcal{M}_v$ & 39.4 \\
+$\epsilon$ & 48.7 \\
+I ($\mathcal{M}_a$) & 49.7\\
\bottomrule[0.1em]
\end{tabular}
\caption{Pseudo masking strategy on LDIS-n.}
\label{tab:ab_pm}
\end{minipage}
\end{table}

\begin{table}[ht]
\centering
\footnotesize
\begin{tabular}{ccccc}
    \toprule[0.1em]
    Meta-architecture & Full Train & Rank & PO & mIoU \\ 
    \midrule[0.1em]
    \multirow{4}{*}{LDIS-1} & - & 1 & - & 53.0 \\
    & - & 4 & - & 53.2 \\ 
    & - & 4 & \checkmark & 55.6 \\ 
    & \checkmark & - & - & 61.3 \\ \hline
    \multirow{2}{*}{LDIS-n} & - & 4 & - & 23.2 \\
    & \checkmark & - & - & 49.7\\
    \bottomrule[0.1em]
    \end{tabular}
\caption{Meta-architecture strategy. PO means pixel space optimizaiton.}
\label{tab:ab_framework}
\end{table}

\begin{figure}[htbp]
	\centering
	\includegraphics[width=0.8\linewidth]{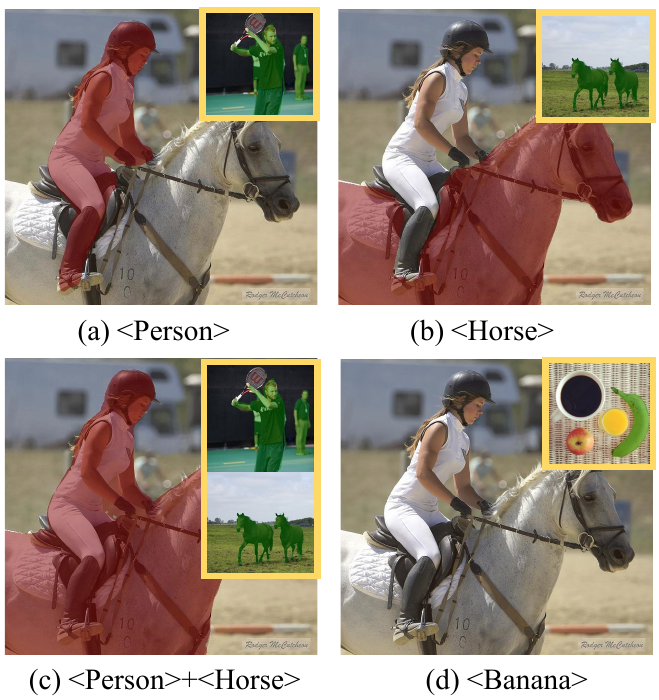}
	\caption{\textbf{Combination of instructions.} The output of our model varies based on the instructions located in the top right corner of the query images. (a) and (b) Single instruction. (c) Multiple instructions. (d) Incorrect instruction.}
	\label{fig:inscomb}
\end{figure}

In addition, we study the combination of instructions and their influences on performance. Fig.~\ref{fig:inscomb} illustrates some examples. Our model accurately segments the target region in the first and second cases based on the single instruction provided. The third case shows the model can accept several instructions without performance degradation. In the fourth case, the provided instruction becomes ineffective when it conflicts with the query.

\noindent
\textbf{Output Alignment.} We study the effects of output alignment from two aspects. 
One of them is the format of the optimization target. We argue that a more challenging learning target benefits the training process and helps the model avoid learning some shortcuts.  
In Tab.~\ref{tab:ab_pm}, LDIS-n can only achieve 39.4 mIoU with vanilla pseudo masks. We then apply a certain perturbation intensity on masks, forcing the model to differentiate the foreground and background from the statistics rather than remembering the constant value. The perturbation follows a uniform distribution and is added to $\mathcal{M}_v$ as Eq.~\ref{equ:pm2}. Surprisingly, the perturbation brings about a significant improvement of 9.3 mIoU. Afterward, we replace the perturbation with the query image and get $\mathcal{M}_a$, which introduces the semantics. In this way, LDIS-n with $\mathcal{M}_a$ achieves a mIoU of 49.7. 
Another factor is the optimization method. In the second and third lines of Tab.~\ref{tab:ab_framework}, LDIS-1 improves 2.4 mIoU when optimized in the pixel space compared to the latent space. As pixel space directly aligns with the segmentation target, optimization methods like `PO' help reduce the transformation error introduced by VAE and achieve better performance.

\noindent
\textbf{Meta-architecture.} We investigate both meta-architectures by applying different training strategies. Generally, the model with all parameters trainable performs best, with LDIS-1 and LDIS-n getting mIoU of 61.3 and 49.7, respectively. Furthermore, we study the effects of parameter-efficient tuning on our model with LoRA. Performance degradation is observed on both architectures, 8.1 on LDIS-1 and 26.5 on LDIS-n, respectively. 
As we further reduce the rank from 4 to 1, LDIS-1's performance slightly degrades, reaching 53.0 mIoU.
This phenomenon indicates that since SD was initially designed for generative tasks, the restriction of expressive power hinders its transfer to segmentation tasks. LDIS-n is more sensitive to this characteristic.

\begin{table}[t]
\setlength{\tabcolsep}{1.7pt}
\centering
\small
\begin{tabular}{lccccc}
\toprule[0.1em]
Method & Fold-0 & Fold-1 & Fold-2 & Fold-3 & Mean \\ 
\toprule[0.1em]
\multicolumn{1}{l|}{RePRI~\cite{boudiaf2021few}} & 32.0 & 38.7 & 32.7 & 33.1 & 34.1 \\
\multicolumn{1}{l|}{BAM~\cite{lang2022learning}} & 43.4 & 50.6 & 47.5 & 43.4 & 46.2 \\
\multicolumn{1}{l|}{FPTrans~\cite{zhang2022feature}} & 44.4 & 48.9 & 50.6 & 44.0 & 47.0 \\
\multicolumn{1}{l|}{PerSAM~\cite{zhang2023personalize}} & 21.8 & 24.1 & 20.8 & 22.6 & 22.3 \\
\hline
\multicolumn{1}{l|}{LDIS-1} & 59.0 & 64.8 & 59.6 & 57.8 & 60.3 \\ 
\toprule[0.1em]
\end{tabular}
\caption{1-shot segmentation on COCO-$20^i$ using mIoU.}
\label{tab:fewshot}
\end{table}

\noindent
\textbf{Results on One-shot Segmentation.} 
We also test our model under the few-shot segmentation setting. As shown in Tab.~\ref{tab:fewshot}, our model achieves decent performance on all folds of COCO-${20^{i}}$ dataset. 
Specifically, it outperforms some recently proposed generalists, such as PerSAM~\cite{zhang2023personalize}, by a remarkable margin.

\section{Conclusion}
\label{sec:con}
For the first time, we explore and unlock the in-context segmentation capability of latent diffusion models. We empirically study the influential factors of an LDM-based segmentation framework, including instruction extraction, output alignment, and meta-architectures, and highlight the importance of precise instructions, direct optimization targets, and expressive power. We also propose an in-context segmentation benchmark and achieve comparable or even better results than specialists or vision foundation models.

\section{Acknowledgements} 
This work was supported by the National Key Research and Development Program of China (No. 2023YFC3807600). This project was supported by NSFC under Grant No. 62472104.

\footnotesize

\bibliography{aaai25}

\end{document}